\documentclass[letterpaper, 10 pt, conference]{ieeeconf}  

\IEEEoverridecommandlockouts                              

\makeatletter
\let\NAT@parse\undefined
\makeatother

\usepackage{graphics} 
\usepackage{epsfig} 
\usepackage{mathptmx} 
\usepackage{times} 
\usepackage{amsmath} 
\usepackage{amssymb}  
\usepackage{algorithm}
\usepackage{algpseudocode}
\usepackage{float} 
\usepackage[dvipsnames]{xcolor}
\usepackage{balance}
\usepackage{url}
\usepackage{cite}
\usepackage{mathrsfs}
\usepackage{svg}
\usepackage{graphicx}
\usepackage{capt-of}
\usepackage[hypertexnames=false]{hyperref}

\title{\LARGE 
\textbf{V\textsuperscript{\bfseries 2}-STRep}: \textbf{V}LM-Grounded \textbf{S}tructured \textbf{T}ask \textbf{Rep}resentations for Reusable Robot Skills Acquired from Generated \textbf{V}ideos
}

\author{Yexin Hu$^{1}$, Dongheui Lee$^{1, 2}$
\thanks{$^{1}$Yexin Hu, and Dongheui Lee are with the Autonomous Systems Lab, Technische Universität Wien (TU Wien), Austria. Emails: \{yexin.hu, dongheui.lee\}@tuwien.ac.at}%
\thanks{$^{2}$Dongheui Lee is with the Institute of Robotics and Mechatronics, German Aerospace Center (DLR), Germany.}%
}

\begin{document}

\maketitle
\thispagestyle{empty}
\pagestyle{empty}

\begin{abstract}
Human manipulation videos provide rich motion and interaction cues for acquiring robot skills without robot demonstrations. Video generation models synthesize such demonstrations from an initial scene image and task instruction, avoiding the need to record demonstrations for each task. However, the recovered motion captures only one scene-specific realization, leaving task structure, geometric relations, and constraints implicit. We present V\textsuperscript{\bfseries 2}-STRep, a zero-shot framework that converts generated video motion into reusable robot skills through VLM-grounded structured task representations. The representation specifies motion phases, references, and task-relevant constraints, with targets described by minimal geometric structures: points, point-normals, axes, planes, and full 6D poses. VLM-provided 2D image-space cues are lifted into 3D using RGB-D observations to reconstruct task geometry and candidate grasp poses. Geometry-specific rules transfer motion to new scenes, while task-constrained trajectory optimization couples grasp selection with complete robot motion planning. It preserves task requirements while using remaining rotational freedom to accommodate joint limits. Updating deployment grounding and constraints enables reuse under new compatible instructions without generating another video. Experiments on six real-world manipulation tasks demonstrate improved execution success over baselines, reliable cross-scene transfer of successfully acquired skills, and adaptation to changed deployment instructions. Details are available on the project website: \href{https://tuwien-asl.github.io/V2-STRep/}{\nolinkurl{https://tuwien-asl.github.io/V2-STRep/}}.

\end{abstract}

\section{Introduction}

Recent advances in foundation models have expanded robot manipulation capabilities in open-world settings~\cite{brohan2023rt,huang2023voxposer,huang2024rekep,kim2024openvla,black2024pi_0}. However, many approaches still rely on substantial robot data, environment-specific adaptation, or predefined skill primitives, limiting scalability across tasks and scenes. Human manipulation videos provide rich motion and interaction cues, and video generation models can synthesize such demonstrations from an initial scene image and task instruction. This has motivated methods that recover object trajectories or 3D flows from generated videos for zero-shot robot execution~\cite{patel2026robotic,li2025novaflow,dharmarajan2025dream2flow}.

\begin{figure}[t]
    \centering
    \includegraphics[width=\columnwidth]{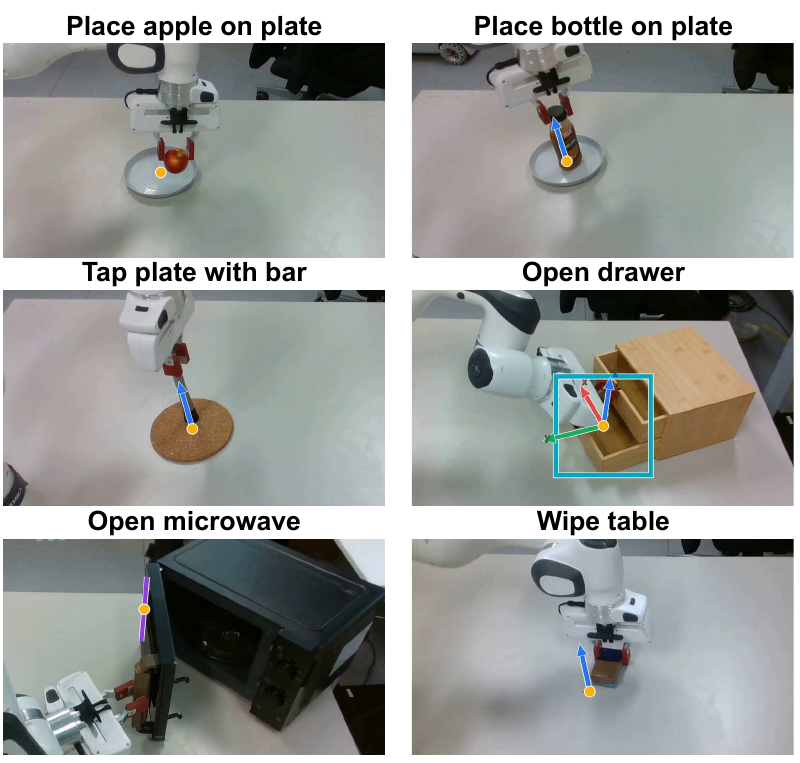}
    \vspace{-0.8cm}
    \caption{
    Six real-world tasks evaluated with V$^{2}$-STRep. Overlays illustrate the VLM-selected target geometries: point (apple), point-normal (bottle and bar),
    full 6D pose (drawer), axis (microwave), and plane (wiping).}
    \vspace{-0.7cm}
    \label{fig:teaser}
    
\end{figure}

However, the recovered trajectory from those methods describes only one scene-specific realization of a task without considering its executability. Recent methods begin to address the executability of recovered trajectories~\cite{huang2026physv2a,huang2026genvid2robot,zhang2026emboalign,he2026roboreact}, while the underlying task structure often remains implicit. A reusable skill should specify which motion is task-critical, what it is referenced to, and which geometric and orientation requirements must be preserved across scenes. For example, upright bottle placement constrains tilt while allowing rotation about the bottle axis, whereas wiping requires sponge--table contact. Distinguishing task requirements from remaining motion freedom allows trajectory optimization to preserve task behavior while improving robot feasibility.

We introduce \textbf{V\textsuperscript{\bfseries 2}-STRep}, a framework that converts generated video motion into a structured task representation for reusable robot skills (Fig.~\ref{fig:pipeline}). Given a video, its initial scene, and the task instruction, a VLM identifies motion phases, references, and task-relevant constraints. Each phase retains video-derived motion or replaces it with a connecting motion to the starting pose of the next task-relevant phase. Retained motion is expressed either relative to the manipulated-object pose at phase entry or to the task target. Orientation constraints specify which aspects of object orientation are constrained and which remain free for optimization. Together, these elements define \emph{which motion to reuse, relative to what, and under which constraints}.

We represent each task target using the minimal geometric structure required for execution: a \emph{point}, \emph{point-normal}, \emph{axis}, \emph{plane}, or \emph{full 6D pose}. Fig.~\ref{fig:teaser} shows the six evaluated manipulation
tasks and their VLM-selected target geometries. These geometries are reconstructed by lifting VLM-selected 2D image cues into 3D using RGB-D observations. Geometry-specific rules then re-instantiate acquired motion with the deployment target geometry. Grasp candidates are similarly reconstructed from VLM-selected pixels and qualitative gripper directions.

Once re-instantiated in the new scene, the transferred task representation is converted into robot motion through trajectory optimization that jointly considers grasp choice, kinematics, smoothness, and collision avoidance. It preserves task requirements while using remaining rotational freedom to accommodate joint limits. Updating deployment grounding and orientation policies further enables adaptation to new compatible instructions under deployment scene while retaining the acquired motion structure, without generating another video.

The contributions of this paper are:
\begin{itemize}
\item \textbf{A structured task representation} for reusable video-derived skills that decomposes motion into phases and specifies which motion to retain or replace, its reference, and the minimum task-relevant geometry and orientation requirements which enables trajectory optimization to preserve task behavior while improve robot flexibility.
\item \textbf{A VLM-guided grounding and geometry-specific motion transfer mechanism} that lifts 2D image-space cues into 3D task geometry and candidate grasp poses using RGB-D observations. The reconstructed geometry determines how the acquired motion is re-instantiated under changes in target position and orientation.
\item \textbf{A zero-shot framework} that acquires reusable robot skills from generated video, transfers to new scenes and adapts to new compatible deployment instructions without generating additional videos.
\end{itemize}

\section{Related Works}
\subsection{Generated Videos for Robot Skill Acquisition}
Generated videos provide task-motion priors for robot manipulation. Earlier approaches typically require train video generators or downstream action models for robot execution~\cite{du2023learning,ko2024learning,bharadhwaj2024gen2act,liang2024dreamitate}. Recent methods instead leverage off-the-shelf video generators at inference time and recover 6D object trajectories~\cite{patel2026robotic} or 3D object flow~\cite{li2025novaflow, dharmarajan2025dream2flow} for zero-shot robot execution. More recent works have focused on the reliability of generated-video execution by incorporating rigid-geometric consistency, VLM-generated compositional constraints or re-grounding mechanisms~\cite{huang2026genvid2robot,zhang2026emboalign,he2026roboreact}. In contrast, V\textsuperscript{2}-STRep acquires directly an task representation that governs both motion transfer across scenes and task-constrained robot execution.

\subsection{Structured and Object-Centric Task Representations}
Structured and object-centric representations support skill transfer across different scene configuration. Some methods represent motion relative to task-relevant objects to adapt it to new spatial arrangement~\cite{heppert2024ditto,li2024okami}, while others represent task geometry through object-centric keypoints, local frames, and movement primitives~\cite{gao2023k}. More recent approaches introduce higher-level structure through object-centric graphs, sparse waypoints, phase decompositions, or reusable keyframes~\cite{zhu2026vision, jonnavittula2025view,he2026roboreact, caccavale2019kinesthetic}. Our representation explicitly couples phase-wise motion type, reference selection, task-dependent target geometry and orientation constraints, which jointly specifies which task behavior to preserve and how it should be re-instantiated in a new scene.

\subsection{VLM-Based Geometric Grounding for Manipulation}

VLM-based manipulation uses intermediate spatial representations to bridge semantic reasoning and robot execution. Some approaches use visual prompts or spatial points to identify task-relevant interaction locations~\cite{nasiriany2024pivot,liu2024moka,yuan2024robopoint,tang2025kalie,zhu2026spatialpoint}. Others represent manipulation through 2D motion paths or contact trajectories that provide coarse guidance for downstream control~\cite{li2025hamster,xu2025a0}. Keypoint-based methods instead encode task objectives through semantic keypoints or relational constraints, which can be converted into planning costs or rewards~\cite{liu2025kuda,zhu2025bridging,patel2025real,huang2024rekep}. More explicit geometric representations describe interactions through object-part relations or geometric constraints~\cite{huang2024copa,tang2025geomanip}, while other approaches construct object-centric interaction primitives or continuous 3D value maps for motion generation~\cite{pan2025omnimanip,huang2023voxposer}.

V\textsuperscript{2}-STRep selects a task-dependent target geometry: \emph{point}, \emph{point-normal}, \emph{axis}, \emph{plane}, or \emph{full 6D pose}, and reconstructs it from VLM-selected 2D cues using RGB-D observations. Each geometry determines the motion-transfer rule, while phase-wise orientation policies specify the rotational degree of freedom for robot optimization and execution.

\section{V\textsuperscript{\bfseries 2}-STRep}

\begin{figure*}[t]
    \centering
    \includegraphics[width=\textwidth]{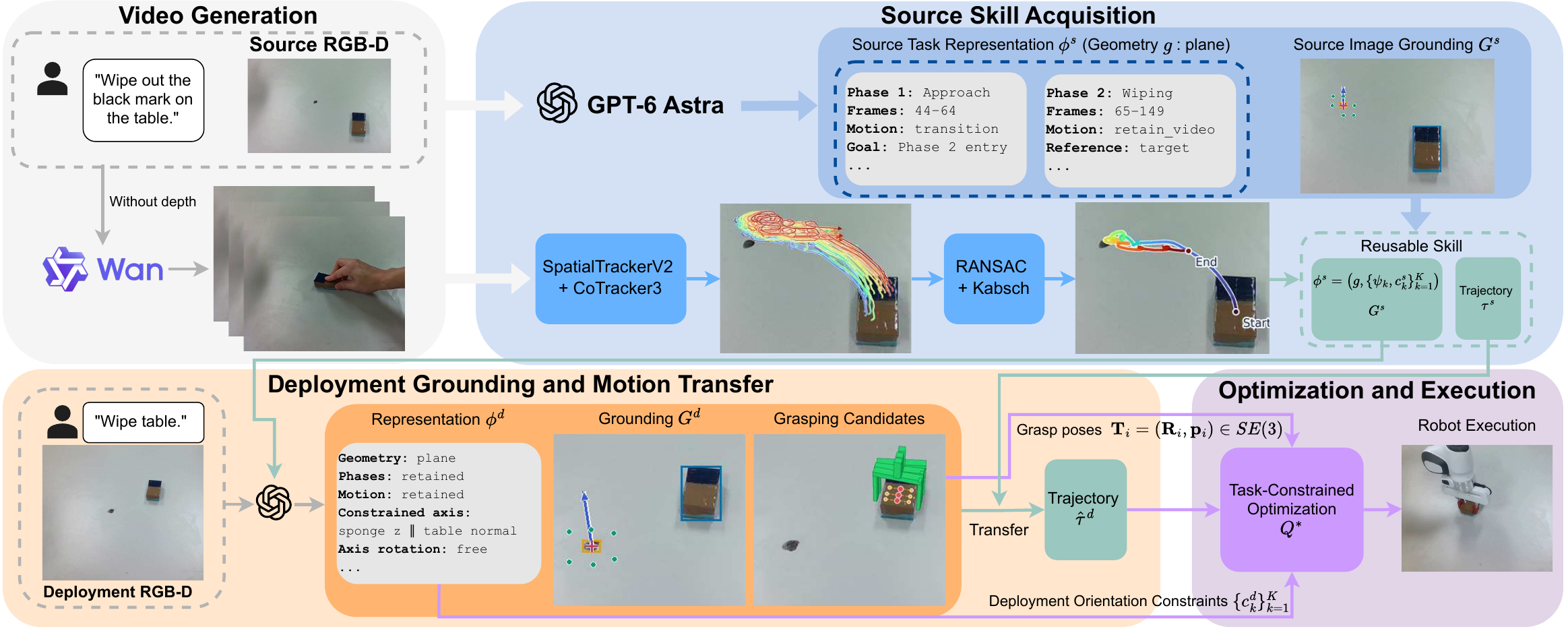}
    \vspace{-0.8cm}
    \caption{Pipeline of V$^{2}$-STRep, illustrated with table wiping task. VLM inference, grounding, and 3D motion recovery produce a reusable skill. Deployment grounding and motion transfer adapt the skill to a new scene. Task-constrained optimization combines grasp selection with complete trajectory planning for robot execution.}
    \vspace{-0.7cm}
    \label{fig:pipeline}
\end{figure*}

V\textsuperscript{\bfseries 2}-STRep combines video-derived object motion with a structured task representation to support skill transfer and robot execution. Fig.~\ref{fig:pipeline} summarizes the overall pipeline.

\subsection{Problem Formulation and Framework Overview}
\label{III.A}

Our goal is to acquire a skill from a source scene and instruction $l^{s}$, and execute it in a deployment scene under a new compatible instruction $l^{d}$. We assume RGB-D observations of both scenes, with RGB images denoted by $I_0^{s}$ and $I_0^{d}$. The acquired skill is reused without additional video generation or robot demonstrations.

Wan 2.7~\cite{wan2025wan} as the video generation model generates a source video $V^{s}$ from $I_0^{s}$ and $l^{s}$. GPT-6 Astra as the VLM, analyzes the image, instruction, and video to infer the task representation $\phi^{s}$ and source image grounding $G^{s}$. The representation specifies motion phases, references, target geometry, and orientation requirements while the grounding identifies the manipulated object and task target through 2D pixels and bounding boxes. In parallel, we recover the source object trajectory $\tau^{s}=\{\mathbf{T}^{s}(t)\}_{t=1}^{N}$ from generated video, where $\mathbf{T}^{s}(t)$ denotes the object pose at frame $t$.

At deployment, the VLM analyzes $I_0^{d}$, $l^{d}$, and $\phi^{s}$ to infer deployment grounding $G^{d}$ and an adapted representation $\phi^{d}$. The source phase decomposition, motion references, and target-geometry type are inherited from $\phi^{s}$, while orientation requirements are re-evaluated for the deployment task. RGB-D grounding reconstructs task geometry in both scenes and candidate grasp poses in the deployment scene. The reconstructed geometry and phase-wise reference determine how $\tau^{s}$ is transferred into the deployment object reference trajectory $\hat{\tau}^{d}$. For each grasp candidate, we optimize a complete joint trajectory that tracks $\hat{\tau}^{d}$ under the deployment task requirements, while using remaining rotational freedom to improve robot feasibility. We select the feasible grasp--trajectory pair with the lowest selection cost, yielding the final joint trajectory $\mathbf{Q}^{*}$.

\subsection{VLM-Inferred Structured Task Representation}
\label{III.B}

We represent the skill at source and deployment scene as
\begin{equation}
\phi^{x}=\left(g,\{\psi_k,c_k^{x}\}_{k=1}^{K}\right),
\quad x\in\{s,d\},
\end{equation}
where $g$ denotes the task target geometry type, $\psi_k$ describes the reusable motion structure of phase $k$, and $c_k^{x}$ specifies its task-relevant orientation constraints. The task representation is inferred by VLM using a vocabulary shared across tasks.

\subsubsection{Target geometry}
Each skill uses a single task target. The VLM selects the simplest geometry type $g$ sufficient for the target-relative motion and orientation requirements of all phases. Thus, $g$ is shared across phases while their references and constraints determine how it is used. The choices are \texttt{point} for a target position used when target-relative orientation is irrelevant; \texttt{point\_normal} for a localized target position and surface normal, e.g. upright placement on a surface; \texttt{axis} captures a physical or functional axis for motion such as opening a hinged door; \texttt{plane} for an extended interaction surface such as a table for wiping; and \texttt{full\_6D\_pose} when a complete target frame is required.

\subsubsection{Phase-wise motion structure}
The VLM infers both $K$ phases and their phase boundaries from the video and instruction. It is prompted to use the fewest, ordered and non-overlapping phases needed to describe the task. Each phase is represented as
\begin{equation}
\psi_k=(w_k,m_k,r_k),
\end{equation}
where $w_k=[t_k^{\mathrm{start}},t_k^{\mathrm{end}}]$ denotes its frame interval. The motion source $m_k$ specifies how its trajectory is obtained:
the \texttt{retain\_video} mode uses the recovered object motion
of that phase as the reference for transfer and optimization; the \texttt{transition} mode replaces it with a connecting motion to the entry pose of the subsequent task-relevant phase.

For \texttt{retain\_video} phases, $r_k$ specifies whether the recovered motion is expressed relative to the object pose at phase entry (\texttt{phase\_initial}) or to the target geometry (\texttt{target}). The reference $r_k$ is unused for \texttt{transition} phases. For example, the table-wiping task in Fig.~\ref{fig:pipeline} consists of an approach phase using \texttt{transition}, followed by a wiping phase using \texttt{retain\_video} with $r_k=\texttt{target}$.

\subsubsection{Phase-wise orientation constraints}
Each phase further specifies an orientation policy
\begin{equation}
c_k^{x}=(o_k^{x},A_k^{x},\rho_k^{x}).
\end{equation}
The mode $o_k^{x}$ determines which aspects of object orientation are constrained:
\begin{itemize}
    \item \texttt{free} imposes no task-specific orientation requirement.
    \item \texttt{axis\_constrained} uses $A_k^{x}$ to specify the controlled object axis, its reference direction, and their required alignment relation. The reference direction comes from the target geometry or an object-axis direction at phase entry. For example, upright bottle placement aligns the bottle's upright axis with the plate normal.
    \item \texttt{full\_tracking} tracks the complete video-derived orientation at each frame, using the transferred reference at deployment.
\end{itemize}

Motion retention is determined separately by $m_k$. For $o_k^{x}=\texttt{axis\_constrained}$, $\rho_k^{x}$ specifies how the remaining rotational freedom is treated: \texttt{free} leaves rotation about the aligned axis to the optimizer, while \texttt{fixed\_from\_video} adds a soft preference toward the complete video-derived orientation alongside the
axis constraint.

At deployment, $g$ and $\{\psi_k\}_{k=1}^{K}$ are retained, while the orientation policies $\{c_k^{d}\}_{k=1}^{K}$ may be updated for the new scene and instruction. Reuse requires compatibility with the acquired target-geometry type and phase structure.

\subsection{Task Geometry and Grasp Grounding}
\label{III.C}

The target geometry $g$ selected in Sec.~\ref{III.B} and candidate grasp poses are reconstructed by lifting VLM-provided 2D image cues into 3D using RGB-D observations.

\subsubsection{VLM-guided geometric grounding}
For each scene $x\in\{s,d\}$, grounding $G^{x}$ contains VLM-selected pixels and bounding boxes identifying the manipulated object and task target. These cues select depth samples and image regions for reconstruction. With depth $D^{x}$ and camera calibration $C^{x}$, they determine the 3D target geometry:
\begin{equation}
\Gamma_g^{x}=\mathcal{C}_g(G^{x},D^{x},C^{x}),
\end{equation}
where $\mathcal{C}_g$ denotes the reconstruction procedure for geometry type $g$, and $\Gamma_g^{x}$ is the resulting 3D target geometry.

For \texttt{point}, the selected target pixel is back-projected to obtain its 3D position. For \texttt{point\_normal} and \texttt{plane}, lifted surface pixels additionally provide a surface normal alongside the target position or plane reference point. For \texttt{axis}, lifted pixels along the axis determine its direction, while an additional reference pixel specifies its origin. For \texttt{full\_6D\_pose}, SAM2~\cite{ravi2025sam} segments the grounded target region in each scene to obtain RGB-D point clouds. The source target frame is initialized from its point cloud. LoFTR~\cite{sun2021loftr} establishes 2D correspondences between the source and deployment images. We lift the matched pixels into 3D using depth and camera calibration, then estimate a rigid transformation through robust registration of the resulting 3D correspondences. Applying this transformation to the source target frame yields the deployment target frame.

The reconstructed target geometry is represented as
\begin{equation}
\Gamma_g^{x}=
\begin{cases}
\mathbf{p}_g^{x},
& g=\texttt{point},\\
(\mathbf{p}_g^{x},\mathbf{n}^{x}),
& g=\texttt{point\_normal},\\
(\mathbf{p}_g^{x},\mathbf{a}^{x}),
& g=\texttt{axis},\\
(\mathbf{p}_g^{x},\mathbf{n}^{x}),
& g=\texttt{plane},\\
\mathbf{T}_g^{x},
& g=\texttt{full\_6D\_pose},
\end{cases}
\end{equation}
where $\mathbf{p}_g^{x}$ denotes the target position, axis origin, or plane reference; $\mathbf{n}^{x}$, $\mathbf{a}^{x}$ are the unit surface normal and axis direction, respectively, and $\mathbf{T}_g^{x}\in SE(3)$ is the full target frame.

For all geometry types, the manipulated object is similarly segmented and registered to obtain consistent initial object frames across scenes as $g$ = \texttt{full\_6D\_pose}.

\subsubsection{VLM-guided grasp grounding}

Given the deployment image and instruction, the VLM generates three grasp proposals, each specifying an anchor pixel, a pointing mode, and an opening-direction cue. The opening direction is the axis along which the gripper fingers move apart or together.

For proposal $i$, the anchor is back-projected to $\mathbf{p}_i$, and neighboring 3D points together provide a local surface normal. The pointing mode sets the unit direction $\mathbf{z}_i$ either downward in the world frame or inward along this normal.

The VLM specifies the opening direction using either the local normal or two pixels across the intended grasp width. The normal option is used only for downward grasps when it provides a suitable closing direction and is not nearly parallel to $\mathbf{z}_i$. All other proposals use the two-pixel option. The VLM-selected cue is reconstructed in 3D, projected perpendicular to $\mathbf{z}_i$, and normalized to obtain $\mathbf{y}_i$. The resulting grasp pose is $(\mathbf{R}_i,\mathbf{p}_i)\in SE(3)$, with
\begin{equation}
\mathbf{R}_i=
\begin{bmatrix}
\mathbf{y}_i\times\mathbf{z}_i &
\mathbf{y}_i &
\mathbf{z}_i
\end{bmatrix}.
\end{equation}
Each pose is augmented with a $180^\circ$ rotation about $\mathbf{z}_i$, yielding up to six poses for grasp selection and trajectory optimization in Sec.~\ref{III.E}.

\subsection{Video Motion Recovery and Transfer}
\label{III.D}

We first recover the manipulated-object motion from the generated video, then transfer it using phase structure in Sec.~\ref{III.B} and the grounded geometry in Sec.~\ref{III.C}.

\subsubsection{Video motion recovery}

Using the source object mask obtained in Sec.~\ref{III.C} , we follow the dense motion recovery pipeline of Dream2Flow~\cite{dharmarajan2025dream2flow}. Depth predictions from SpatialTrackerV2~\cite{xiao2025spatialtrackerv2,yang2024depth} and 2D tracks from CoTracker3~\cite{karaev2025cotracker3} are combined to recover 3D object flows. We estimate rigid object motion relative to a common reference using RANSAC-based registration~\cite{fischler1981random} followed by Kabsch refinement and temporal smoothing, yielding the source rigid-body trajectory $\tau^{s}=\{\mathbf{T}^{s}(t)\}_{t=1}^{N}$.

\subsubsection{Geometry-conditioned motion transfer}
For \texttt{retain\_video} phases, the recovered object motion provides a reference that is transferred relative to $r_k$. For \texttt{transition} phases, the video-derived path is replaced by a connecting motion to next \texttt{retain\_video} phase.

\paragraph{Phase-initial reference}
For $r_k=\texttt{phase\_initial}$, motion is preserved relative to the object pose at phase entry. Let $t_k^{\mathrm{start}}$ denote the first frame of phase $k$. We compute and transfer the relative motion as
\begin{equation}
\begin{aligned}
\Delta\mathbf{T}_k^{s}(t)
&= \mathbf{T}^{s}(t_k^{\mathrm{start}})^{-1} \mathbf{T}^{s}(t),\\
\hat{\mathbf{T}}_k^{d}(t)
&=\hat{\mathbf{T}}_k^{d}(t_k^{\mathrm{start}}) \Delta\mathbf{T}_k^{s}(t),
\end{aligned}
\end{equation}
where $\hat{\mathbf{T}}_k^{d}(t_k^{\mathrm{start}})$ is the deployment phase-entry pose.

\paragraph{Target reference}

For $r_k=\texttt{target}$, motion is transferred relative to the target geometry. Writing the source object pose as
$\mathbf{T}^{s}(t)=(\mathbf{R}^{s}(t),\mathbf{p}^{s}(t))$, we obtain
\begin{equation}
\begin{aligned}
\hat{\mathbf{R}}_k^{d}(t)
&=\mathbf{R}_g^{s\rightarrow d}\mathbf{R}^{s}(t),\\
\hat{\mathbf{p}}_k^{d}(t)
&=\mathbf{p}_g^{d}+\mathbf{R}_g^{s\rightarrow d}\mathbf{P}_g^{s}\left(\mathbf{p}^{s}(t)-\mathbf{p}_g^{s}\right),
\end{aligned}
\label{eq:target_transfer}
\end{equation}
where $\mathbf{p}_g^{x}$ is the target reference position defined in Sec.~\ref{III.C}. The rotation $\mathbf{R}_g^{s\rightarrow d}$ aligns the task-relevant source and deployment geometry, while $\mathbf{P}_g^{s}$ selects the translational components to retain. We use $\mathbf{P}_g^{s}=\mathbf{I}$ except for \texttt{plane}.

For \texttt{point}, no target-relative rotational alignment is required. We set
$\mathbf{R}_{\texttt{point}}^{s\rightarrow d}=\mathbf{I}$,
translating the reference trajectory with respect to the target while retaining its video-derived orientation reference.

For \texttt{point\_normal}, let $\mathbf{n}^{s}$ and $\mathbf{n}^{d}$ denote the source and deployment surface normals. We set $\mathbf{R}_{\texttt{point\_normal}}^{s\rightarrow d}=\mathbf{R}_{\min}(\mathbf{n}^{s},\mathbf{n}^{d})$,
where $\mathbf{R}_{\min}$ denotes the minimum-angle rotation aligning the two surface normals. This adapts the motion reference to the changed surface normal without introducing an additional rotation about the normal.

For \texttt{plane}, the same normal alignment is used, while the displacement normal to the source plane is removed:
\begin{equation}
\mathbf{R}_{\texttt{plane}}^{s\rightarrow d}=\mathbf{R}_{\min}(\mathbf{n}^{s},\mathbf{n}^{d}),\
\mathbf{P}_{\texttt{plane}}^{s}=\mathbf{I}-\mathbf{n}^{s}{\mathbf{n}^{s}}^{\top}.
\end{equation}
This preserves the in-plane motion while adapting it to the orientation and position of the deployment plane.

For \texttt{axis}, we first align the axis directions using
$\mathbf{R}_0=\mathbf{R}_{\min}(\mathbf{a}^{s},\mathbf{a}^{d})$.
An additional rotation about the deployment axis accounts for changes in the object's initial angular position around this axis. We determine this rotation using the initial object positions $\mathbf{p}_0^{s}$ and $\mathbf{p}_0^{d}$, whose radial vectors after axis alignment are
\begin{equation}
\begin{aligned}
\mathbf{r}^{s}
&=\left(\mathbf{I}-\mathbf{a}^{d}{\mathbf{a}^{d}}^{\top}\right)\mathbf{R}_0\left(\mathbf{p}_0^{s}-\mathbf{p}_g^{s}\right),\\
\mathbf{r}^{d}
&=\left(\mathbf{I}-\mathbf{a}^{d}{\mathbf{a}^{d}}^{\top}\right)\left(\mathbf{p}_0^{d}-\mathbf{p}_g^{d}\right).
\end{aligned}
\end{equation}
The signed angle from $\mathbf{r}^{s}$ to $\mathbf{r}^{d}$ about $\mathbf{a}^{d}$ determines the additional rotation:
\begin{equation}
\begin{aligned}
&\theta=\operatorname{atan2}\!\left({\mathbf{a}^{d}}^{\top}(\mathbf{r}^{s}\times\mathbf{r}^{d}),
{\mathbf{r}^{s}}^{\top}\mathbf{r}^{d}\right),\\
&\mathbf{R}_{\texttt{axis}}^{s\rightarrow d}=\mathbf{R}(\mathbf{a}^{d},\theta)\mathbf{R}_0,
\end{aligned}
\end{equation}
where $\mathbf{R}(\mathbf{a}^{d},\theta)$ denotes rotation by $\theta$ about $\mathbf{a}^{d}$. Thus, $\mathbf{R}_0$ aligns the axis directions, while rotation by $\theta$ aligns the initial radial directions, adapting the video-derived motion to the object's initial position around the deployment axis.

For \texttt{full\_6D\_pose}, the complete target frames are available and the full target-relative transform is preserved:
\begin{equation}
\hat{\mathbf{T}}_k^{d}(t)=
\mathbf{T}_g^{d}
(\mathbf{T}_g^{s})^{-1}
\mathbf{T}^{s}(t).
\end{equation}
which corresponds to
$\mathbf{R}_{\texttt{full\_6D\_pose}}^{s\rightarrow d}=\mathbf{R}_g^{d}(\mathbf{R}_g^{s})^{\top}$
in Eq.~\eqref{eq:target_transfer}.

\paragraph{transition motion source}
For $m_k=\texttt{transition}$, we generate a connecting trajectory from the last phase's end pose to the entry pose of a subsequent \texttt{retain\_video} phase. If \texttt{transition} phase is the first phase, it starts from the initial manipulated object pose.

After all phases are instantiated and concatenated, they form the deployment reference trajectory $\hat{\tau}^{d}$. This reference trajectory is then combined with the phase-wise orientation constraints $c_k^{d}$ and robot-feasibility requirements for trajectory optimization in Sec.~\ref{III.E}.

\subsection{Task-Constrained Robot Trajectory Optimization}
\label{III.E}

We combine grasp selection and trajectory optimization to realize the transferred object motion $\hat{\tau}^{d}$, preserving task requirements while using available rotational freedom to improve robot feasibility.

\subsubsection{Grasp selection and trajectory optimization}

For each grasp candidate $i$ from Sec.~\ref{III.C}, we assume a rigid, no-slip attachment between the gripper and manipulated object. Using PyRoKi~\cite{kim2025pyroki}, we optimize the complete joint trajectory $\mathbf{Q}=\{\mathbf{q}(t)\}_{t=1}^{T}$ under robot-feasibility constraints:
\begin{equation}
\mathbf{Q}_i^{*}=\arg\min_{\mathbf{Q}}\lambda_p E_{\mathrm{pos}}
+\lambda_o E_{\mathrm{ori}}+E_{\mathrm{robot}},
\end{equation}
where $\lambda_p$ and $\lambda_o$ weight position and orientation tracking. The initial joint configuration remains optimizable subject to realizing the grasp pose. The positional objective is
\begin{equation}
E_{\mathrm{pos}}=\sum_{k=1}^{K}\sum_{t\in w_k}
\lambda_{m,k}\|\mathbf{p}_{O}(\mathbf{q}(t))-\hat{\mathbf{p}}^{d}(t)\|_2^2,
\end{equation}
where $\mathbf{p}_{O}(\mathbf{q}(t))$ is the object position obtained through forward kinematics. We use $\lambda_{m,k}=1$ for \texttt{retain\_video} phases and $\lambda_{m,k}=0.03$ for \texttt{transition} phases, giving generated transitions greater flexibility to satisfy robot feasibility. $E_{\mathrm{robot}}$ denotes standard robot feasibility terms, including joint position and velocity limits, trajectory smoothness, acceleration regularization and self-collision penalties. Among feasible grasp--trajectory pairs, we select the one with the lowest selection cost and denote its joint trajectory as $\mathbf{Q}^{*}$.

\subsubsection{Phase-wise task constraints}
The orientation objective implements the phase-wise orientation constraints
$c_k^{d}=(o_k^{d},A_k^{d},\rho_k^{d})$ introduced in Sec.~\ref{III.B}:
\begin{equation}
E_{\mathrm{ori}}=\sum_{k=1}^{K}\sum_{t\in w_k}
\lambda_{m,k}e_{\mathrm{ori}}\left(\mathbf{R}_{O}(\mathbf{q}(t)),c_k^{d}\right),
\end{equation}
where $\mathbf{R}_{O}(\mathbf{q}(t))$ denotes the attached object's orientation. The phase-wise orientation cost is
\begin{equation}
e_{\mathrm{ori}}=
\begin{cases}
0,& o_k^{d}=\texttt{free},\\
\beta_A e_A+\delta_k\beta_R e_F,& o_k^{d}=\texttt{axis\_constrained},\\
\beta_F e_F,& o_k^{d}=\texttt{full\_tracking},
\end{cases}
\end{equation}
where $\beta_A$, $\beta_R$, and $\beta_F$ weight axis alignment, additional orientation regularization, and full-orientation tracking, respectively, and $\delta_k=\mathbb{I}[\rho_k^d=\texttt{fixed\_from\_video}]$.

For \texttt{axis\_constrained}, the relation in $A_k^d$ specifies a controlled unit axis $\mathbf e$ in the object frame and a unit reference direction $\mathbf v$ in the world frame. The reference comes from either the deployment target geometry or the object orientation at phase entry. For alignment with this reference,
\begin{equation}
e_A=\left\|\mathbf{R}_O(\mathbf{q}(t))\mathbf e-\mathbf v\right\|_2^2.
\end{equation}
This penalizes axis misalignment while leaving rotation about the aligned axis free.

The full-orientation error uses the squared Frobenius norm of the rotation-matrix difference:
\begin{equation}
e_F=\left\|\mathbf{R}_O(\mathbf{q}(t))-\hat{\mathbf R}^d(t)\right\|_F^2,
\end{equation}
where $\hat{\mathbf R}^d(t)$ is the transferred orientation reference. For \texttt{axis\_constrained}, $\rho_k^d=\texttt{fixed\_from\_video}$ adds this error as a soft preference toward the complete transferred orientation. When $\rho_k^d=\texttt{free}$, this term is omitted.

Together, task-constrained trajectory optimization and grasp selection yield the final robot joint trajectory $\mathbf{Q}^{*}$.

\vspace{-0.1cm}
\section{Experiments}
\subsection{Experimental Setup}
\label{IV.A}

We conduct real-world experiments using a Franka Emika Panda robot equipped with a parallel-jaw gripper and an Intel RealSense D435i RGB-D camera. The camera provides the initial RGB image for video generation and the RGB-D observations used for task geometry and grasp grounding. 

We use three task-independent VLM prompt templates unchanged across all six tasks: one infers the source representation and grounding; one adapts deployment orientation policies and task grounding while preserving the source phase structure and target geometry; and one proposes three deployment grasp candidates in image space. Together, the deployment templates provide the grounding used in Sec.~\ref{III.C}. The templates specify output schemas and general reasoning rules, without task-specific examples or prescribed solutions. Task-specific information comes from observations, instructions, and the inferred source representation.

The six tasks cover all five target-geometry types in Sec.~\ref{III.B}. The labels below are inferred by the VLM during source-skill acquisition. Success criteria are applied consistently across methods.

\subsubsection{Apple placement (\texttt{point})}
The apple remains on the plate after gripper release.

\subsubsection{Bottle placement (\texttt{point\_normal})}
The bottle remains standing upright on the plate after gripper release.

\subsubsection{Bar tapping (\texttt{point\_normal})}
The bar performs multiple distinct tapping motions, with its tip reaching within 1 cm vertically of the wooden plate surface on each tap, and remains standing upright after gripper release.

\subsubsection{Drawer matching (\texttt{full\_6D\_pose})}
The robot opens the upper drawer to match the lower drawer with its extension to exceed two-thirds of the lower drawer's extension.

\subsubsection{Microwave opening (\texttt{axis})}
The door's opening rotation exceeds two-thirds of that observed in the source video.

\subsubsection{Table wiping (\texttt{plane})}
The sponge visibly wipes away part of the mark drawn on the table during the wiping motion.

We compare with NovaFlow~\cite{li2025novaflow}, which reconstructs rigid object motion from 3D flow for trajectory optimization, and Dream2Flow~\cite{dharmarajan2025dream2flow}, which optimizes robot trajectories to track 3D object flow. Five visually plausible generated videos per task are shared across all three methods (90 trials), with deployment reset to the source configuration. All methods receive the same feasible grasping joint configuration per case, bypassing grasp generation and selection. Recent related methods~\cite{huang2026physv2a,huang2026genvid2robot,zhang2026emboalign,he2026roboreact} are excluded because we could not identify public implementations at evaluation time.

Cross-scene evaluation assumes successful source-skill acquisition: a visually plausible video yielding usable 3D object flow and a valid VLM-inferred task representation. We generate one additional source video per task, extract its flow and representation once, and test the acquired skill across five deployment configurations (30 trials), using our grasp grounding and selection procedure (Secs.~\ref{III.C}--\ref{III.E}). This evaluates the reliability of transferring an already acquired skill under scene changes.

We also evaluate instruction-conditioned reuse qualitatively in three cases (Sec.~\ref{IV.D}).

\vspace{-0.1cm}
\subsection{Comparison with baselines}
\label{IV.B}

With shared source videos and provided grasp configurations, V$^{2}$-STRep succeeds in 25/30 trials, compared with 12/30 for NovaFlow and 8/30 for Dream2Flow (Table~\ref{tab:baseline_comparison}). Our task constraints preserve upright orientations for bottle placement and bar tapping, while the plane-based wiping reference promotes surface contact. Bottle placement succeeds in all five trials while baseline failures include planning failures at joint limits and tilted placements that topple after gripper release. Preserving task-relevant requirements while leaving remaining rotational freedom gives the optimizer greater flexibility to accommodate robot joint limits.

NovaFlow suffers tracking failures in apple placement and drawer matching. Fast motion and hand occlusion destabilize tapping trajectories for all three methods, with our method achieving two successes and neither baseline succeeding. Depth inaccuracies cause downward drift in our failed drawer trial; Dream2Flow exhibits greater drift and succeeds only once. During wiping, inaccurate depth prevents sufficient table contact in failed baseline trials, whereas our single failure involves an ambiguous generated video whose recovered motion does not reach the mark. For microwave opening, Dream2Flow's trajectory optimization produces stationary motion due to its jointly optimization on robot joints and massive tracking points.

\begin{table}[t]
\centering
\caption{Task success rates in the baseline comparison.}
\vspace{-0.3cm}
\label{tab:baseline_comparison}
\small
\setlength{\tabcolsep}{2pt}
\renewcommand{\arraystretch}{1.15}
\begin{tabular}{@{}lccccccc@{}}
\hline
\textbf{Method} &
\textbf{Apple} &
\textbf{Bottle} &
\textbf{Bar} &
\textbf{Drawer} &
\textbf{Micro} &
\textbf{Table} &
\textbf{Overall} \\
\hline
\textbf{NovaFlow}   & 3/5 & 3/5 & 0/5 & 0/5 & \textbf{5/5} & 1/5 & 12/30 \\
\textbf{Dream2Flow} & \textbf{5/5} & 2/5 & 0/5 & 1/5 & 0/5 & 0/5 & 8/30 \\
\textbf{Ours}       & \textbf{5/5} & \textbf{5/5} & \textbf{2/5} & \textbf{4/5} & \textbf{5/5} & \textbf{4/5} & \textbf{25/30}\\
\hline
\end{tabular}
\vspace{-0.5cm}
\end{table}

\vspace{-0.1cm}
\subsection{Cross-Scene Skill Transfer}
\label{IV.C}

\begin{figure}[t]
    \centering
    \includegraphics[width=\columnwidth]{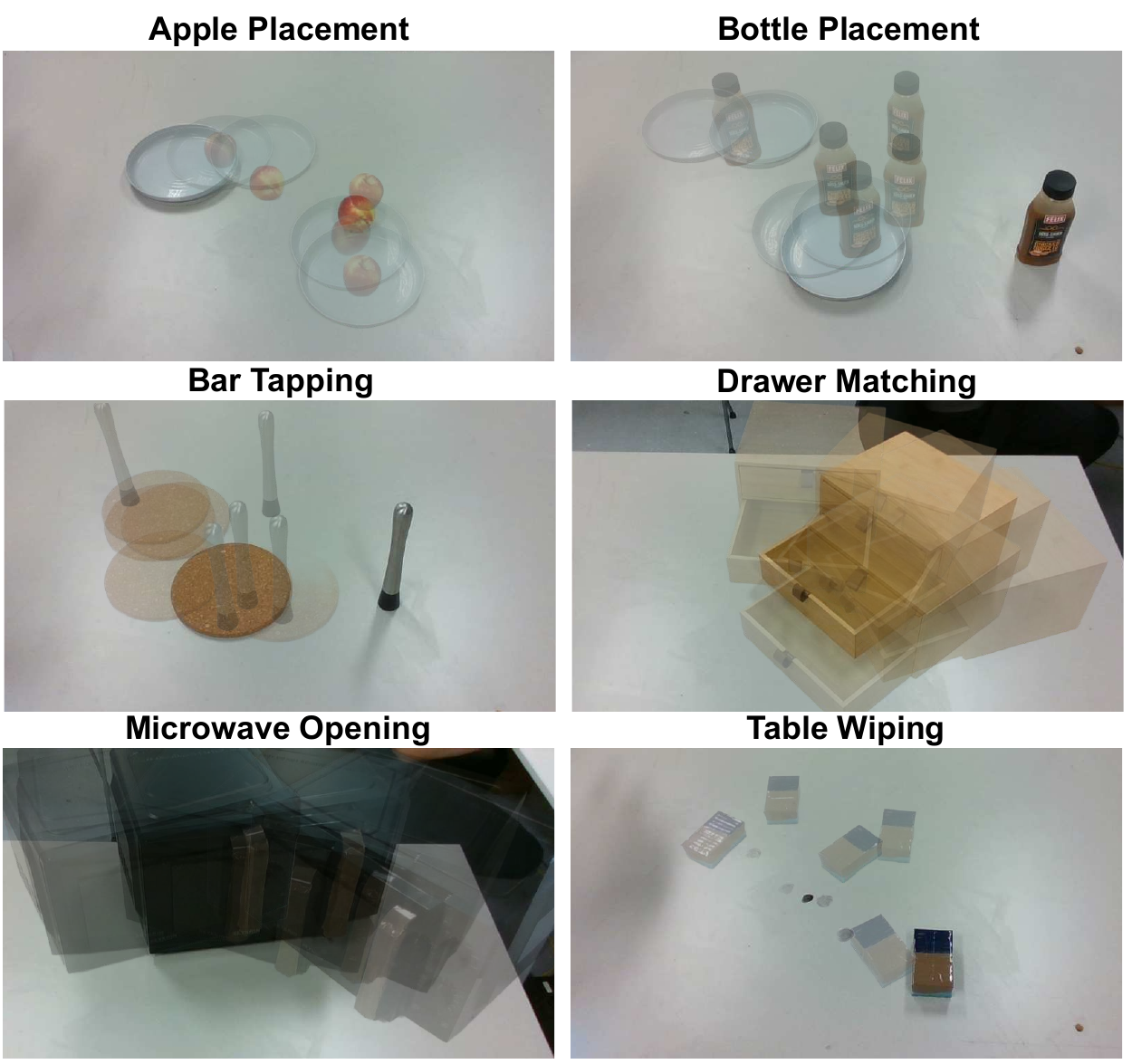}
    \vspace{-0.8cm}
    \caption{Initial object poses for the six tasks, with each source pose shown at full opacity and its five deployment poses overlaid at 30\% opacity.}
    \vspace{-0.7cm}
    \label{fig:cross_scene_configurations}
\end{figure}

\begin{figure*}[t]
    \centering
    \includegraphics[width=\textwidth]{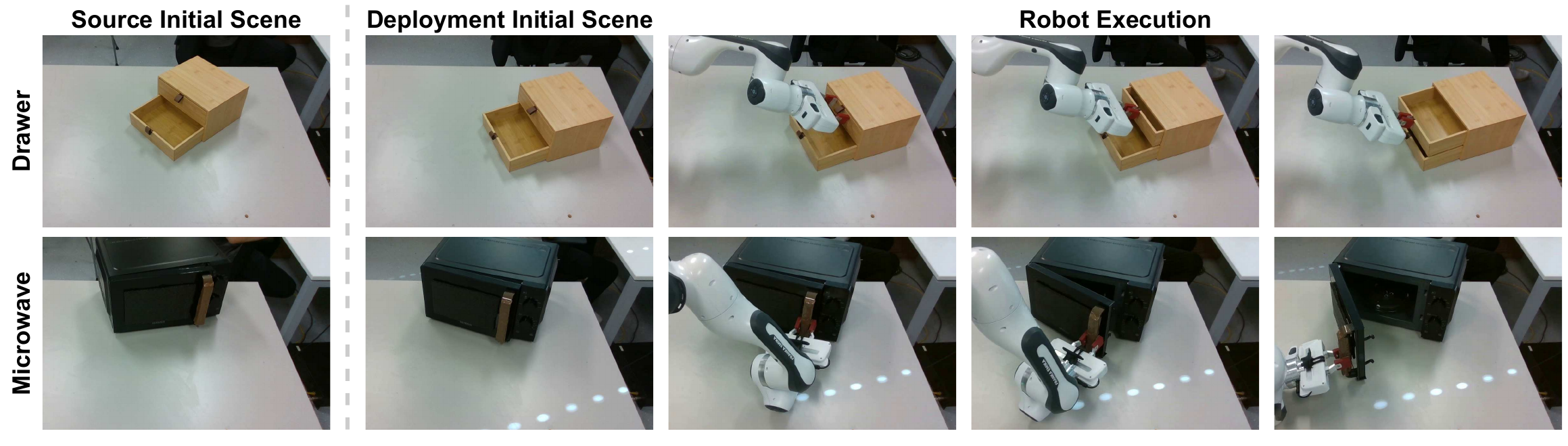}
    \vspace{-0.8cm}
    \caption{Cross-scene transfer for drawer matching and microwave opening, showing source and deployment scenes alongside robot execution.}
    \vspace{-0.5cm}
    \label{fig:cross_scene_execution}
\end{figure*}

Fig.~\ref{fig:cross_scene_configurations} shows the source scenes and five deployment object poses per task. Using our grasp grounding and selection, the method succeeds in 29/30 trials (96.7\%; Table~\ref{tab:cross_scene_transfer}). Fig.~\ref{fig:cross_scene_execution} shows representative successful executions of drawer matching and microwave opening.

The only failure occurs during grasping in table wiping, with the sponge near the image's upper-left corner. Despite visually reasonable VLM cues, inaccurate measured depth reconstructs the sponge point cloud almost flat on the table, placing the grasp anchors on the table surface. These results indicate reliable VLM-guided grounding and motion transfer across the tested configurations given successfully acquired source skills.

\vspace{-0.1cm}
\subsection{Instruction-Conditioned Skill Reuse}
\label{IV.D}

We qualitatively evaluate skill reuse under changed deployment instructions without generating new source videos. The acquired motion structure is retained, while the VLM adapts deployment grounding and orientation constraints.

\subsubsection{Manipulated-object selection}
The apple-placement skill is reused to place a lemon on the plate, with the apple remaining in the scene as a distractor.

\subsubsection{Interaction-target selection}
The table-wiping skill is reused with two visible marks. The changed instruction selects which mark to wipe with acquired skill.

\subsubsection{Orientation-constraint relaxation}
The deployment instruction relaxes the inferred requirement for upright placement, allowing additional rotational freedom during optimization. Success requires the bottle to remain on the plate after release, irrespective of its final orientation.

\begin{figure*}[t]
    \centering
    \includegraphics[width=\textwidth]{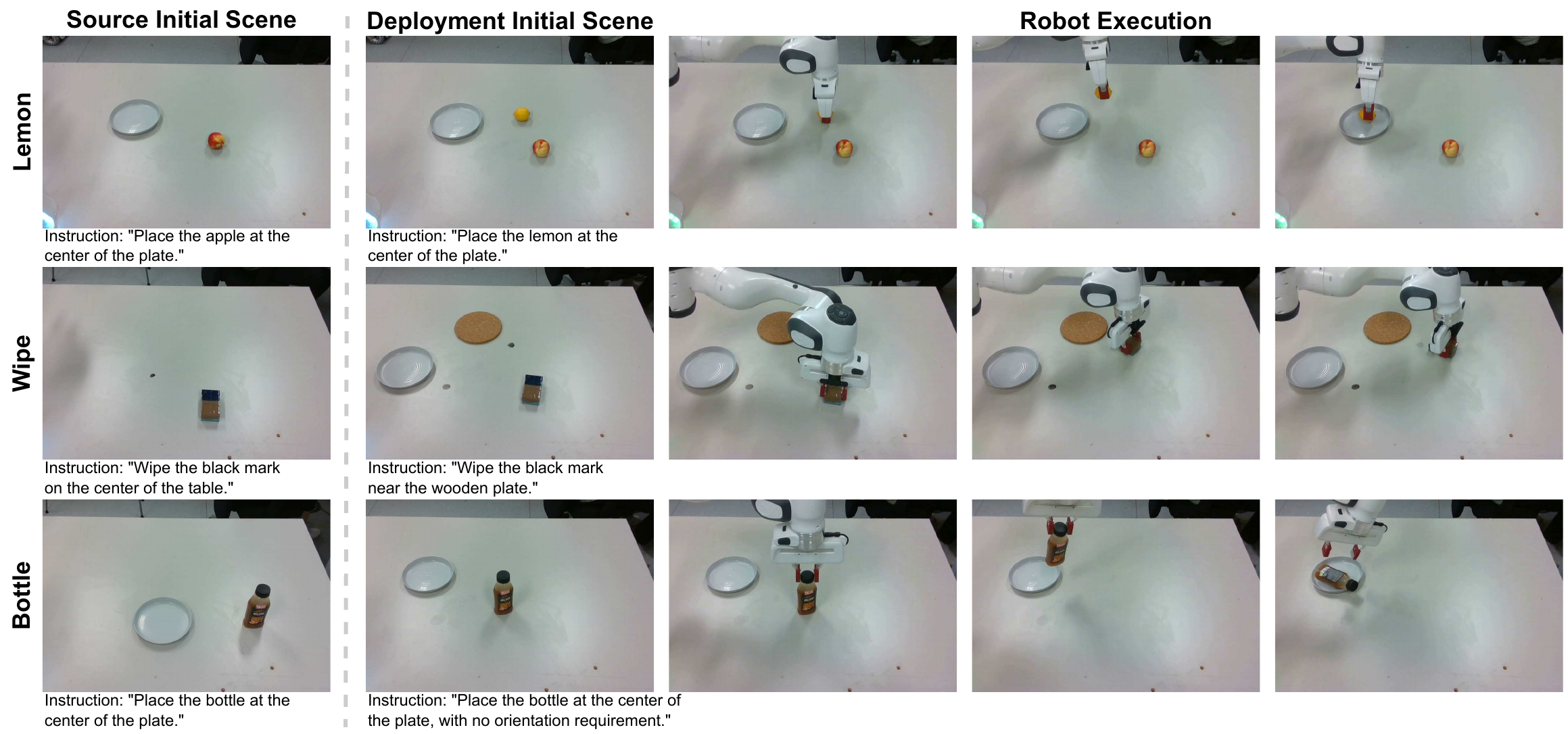}
    \vspace{-0.8cm}
    \caption{Instruction-conditioned skill reuse without new source videos. From top to bottom, deployment instructions change the manipulated object from apple to lemon, select the wiping mark near the wooden plate, and relax the inferred orientation requirement for bottle placement. Each row shows the source scene, deployment scene, and representative execution frames. Instructions are abbreviated for readability.}
    \vspace{-0.8cm}
    \label{fig:instruction_reuse}
\end{figure*}

Fig.~\ref{fig:instruction_reuse} presents the source and deployment instructions alongside representative executions.

\begin{table}[t]
\centering
\caption{Task success rates in cross-scene transfer.}
\vspace{-0.3cm}
\label{tab:cross_scene_transfer}
\small
\setlength{\tabcolsep}{2pt}
\renewcommand{\arraystretch}{1.15}
\begin{tabular}{@{}lccccccc@{}}
\hline
&
\textbf{Apple} &
\textbf{Bottle} &
\textbf{Bar} &
\textbf{Drawer} &
\textbf{Micro} &
\textbf{Table} &
\textbf{Overall} \\
\hline
\textbf{Success rate} &
5/5 & 5/5 & 5/5 & 5/5 & 5/5 & 4/5 & \textbf{29/30} \\
\hline
\end{tabular}
\vspace{-0.7cm}
\end{table}

\section{Conclusions}

We presented V$^{2}$-STRep, a framework for acquiring reusable manipulation skills from generated videos. Its structured representation captures phase-wise motion sources and references, task-relevant target geometry, and orientation requirements, supporting transfer to new scenes and deployment instructions. VLM-provided 2D image cues are lifted into 3D with RGB-D observations to reconstruct task geometry and candidate grasp poses. By optimizing the complete robot trajectory for each grasp candidate, the framework couples grasp selection with task execution, preserving task-relevant behavior while using remaining rotational freedom to improve robot feasibility. Experiments on six real-world manipulation tasks demonstrate improved success over baselines NovaFlow and Dream2Flow, a 96.7\% cross-scene success rate given successfully acquired source skills, and adaptation to changed deployment instructions.

\bibliographystyle{IEEEtran}
\bibliography{IEEEabrv,references}

\end{document}